\documentclass{article}

\usepackage[preprint]{neurips_2024}

\usepackage[utf8]{inputenc} 
\usepackage[T1]{fontenc}    
\usepackage{hyperref}       
\usepackage{url}            
\usepackage{booktabs}       
\usepackage{amsfonts}       
\usepackage{nicefrac}       
\usepackage{microtype}      
\usepackage{xcolor}         
\usepackage{graphicx}
\definecolor{bluebg}{rgb}{.85, .85, 1.0}
\usepackage{caption}
\usepackage{footnote}
\makesavenoteenv{tabular}
\makesavenoteenv{table}
\usepackage{rotating}
\usepackage{tabularray}
\usepackage{fontawesome}

\usepackage{colortbl}

\title{Curating Grounded Synthetic Data with Global Perspectives for Equitable AI}

\author{%
  Elin Törnquist \\
	Emergent Methods, CO\\
	\texttt{elin@emergentmethods.ai} \\
 \And
	Robert Alexander Caulk \\
	Emergent Methods, CO\\
	\texttt{rob@emergentmethods.ai}
}

\begin{document}

\maketitle

\begin{abstract}
  The development of robust AI models relies heavily on the quality and variety of training data available. In fields where data scarcity is prevalent, synthetic data generation offers a vital solution. In this paper, we introduce a novel approach to creating synthetic datasets, grounded in real-world diversity and enriched through strategic diversification. We synthesize data using a comprehensive collection of news articles spanning 12 languages and originating from 125 countries, to ensure a breadth of linguistic and cultural representations. Through enforced topic diversification, translation, and summarization, the resulting dataset accurately mirrors real-world complexities and addresses the issue of underrepresentation in traditional datasets. This methodology, applied initially to Named Entity Recognition (NER), serves as a model for numerous AI disciplines where data diversification is critical for generalizability. Preliminary results demonstrate substantial improvements in performance on traditional NER benchmarks, by up to 7.3\%, highlighting the effectiveness of our synthetic data in mimicking the rich, varied nuances of global data sources. This paper outlines the strategies employed for synthesizing diverse datasets and provides such a curated dataset for NER.
\end{abstract}

\section{Introduction}

In the rapidly evolving field of artificial intelligence, the adequacy of training data significantly influences the accuracy and fairness of AI models \citep{mehrabi2021survey}. Traditionally, datasets have skewed towards predominately available contexts and languages, leading to models that perform inconsistently across diverse global scenarios \citet{osoba2017intelligence}. Recognizing the limitations posed by non-representative datasets, there is an urgent need for a methodological shift towards the creation of more inclusive and diverse datasets. Synthetic data generation emerges as a powerful strategy to address this need, by artificially constructing data that reflects a wider array of linguistic and cultural backgrounds. By grounding synthetic datasets in real-world diversity and enriching them through strategic diversification, we can significantly enhance the performance and generalizability of AI models across different languages and cultures.

News texts offer an exceptional foundation for grounding synthetic data due to their inherent diversity and comprehensive coverage of a multitude of topics and cultures. As dynamic reflections of global events, news articles encapsulate a wide array of subjects, from politics and economy to science and culture, each embedded with nuanced linguistic and contextual specifics characteristic of different geographical regions. This variety makes news texts the ideal base for synthesizing datasets that aim to mirror real-world complexities and societal variations. The recent maturation of Large Language Models (LLMs), such as OpenAI's GPT models \citep{brown2020language, achiam2023gpt}, Meta's Llama models \citep{touvron2023llama, touvron2023llama2}, and Anthropic's Claude models, makes it possible to generate high-quality synthetic text data \citep{ziems2024can} using news articles to incorporate rich, multifaceted inputs that enhance the representativeness and authenticity of the resulting dataset. 

Named Entity Recognition (NER), a fundamental task in natural language processing (NLP) that involves identifying and classifying key elements in text into predefined categories, is a prime example of a field where dataset diversity is a key necessity for off-the-shelf performance. NER systems are integral for a variety of applications, including information retrieval, recommendation systems, and automated content analysis, making their reliability critically important. 
Traditional NER models, such as spaCy \citep{honnibal2020spacy} and Flair \citep{akbik2019flair}, are efficient but limited to a predefined set of entity types. LLMs make it possible to identify any entity type using simple natural language instructions. However, this comes at a significant increase in computational cost. The release of GLiNER \citep{zaratiana2023gliner} bridges the gap between computationally expensive yet entity type permissive LLMs and traditional, restrictive NER frameworks. As opposed to other LLM-based NER models, such as InstructUIE \citep{wang2023instructuie}, GoLLIE \citep{sainz2023gollie}, and UniversalNER \citep{zhou2023universalner}, that use autoregressive language models (LMs), GLiNER uses computationally efficient Bidirectional Language Models (BiLM), such as BERT \citep{devlin2018bert} and deBERTa \cite{he2021debertav3}. This allows GLiNER to predict multiple entity types in parallel, with low computational cost, and permits bidirectional context processing, resulting in richer representations. 

\begin{figure}[h!]
    \centering
    \includegraphics[width=\textwidth]{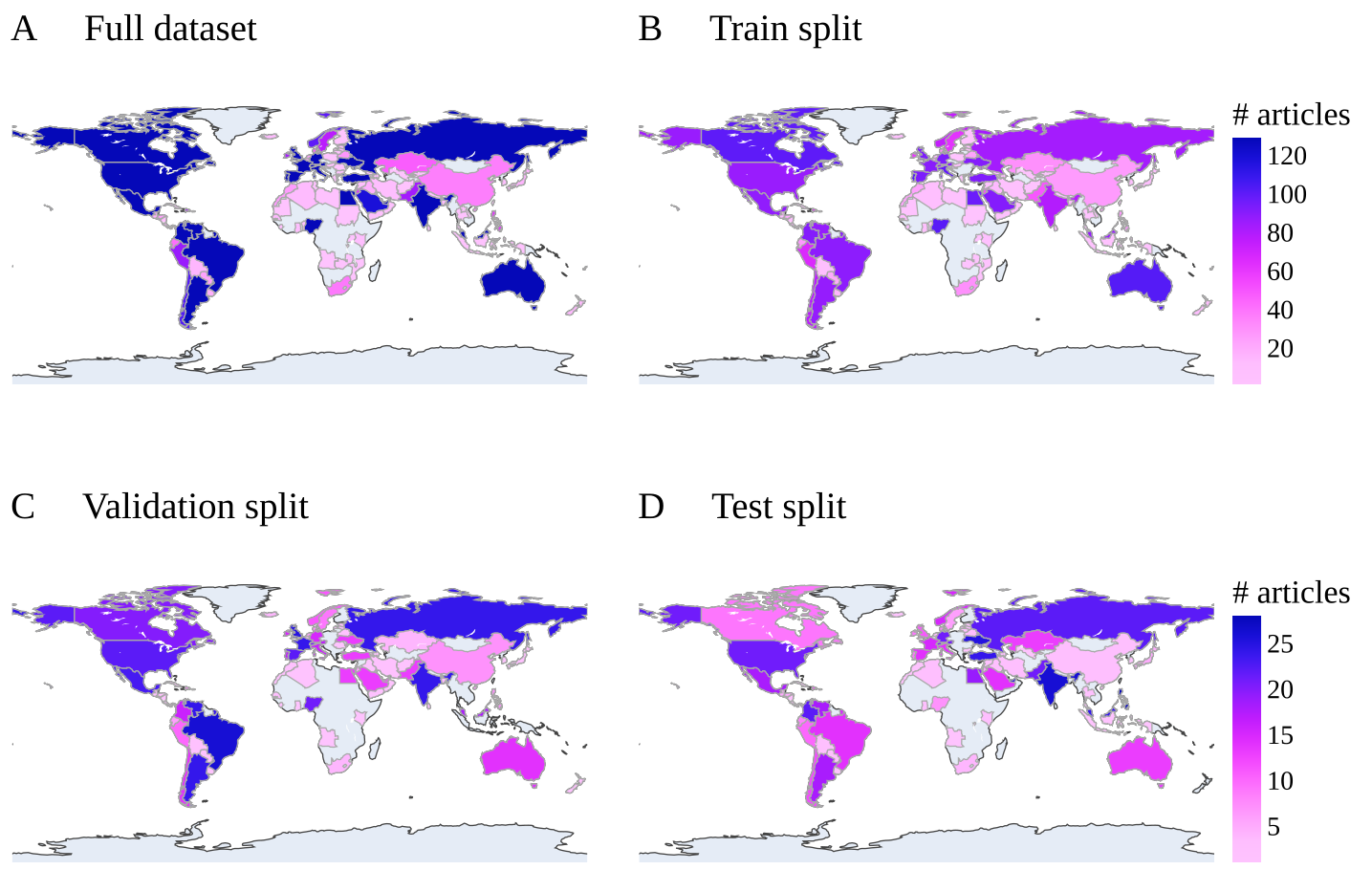}
    \caption{\textbf{Country distribution.} Distribution of country of origin for the articles in the full dataset (A) and in each of the splits (B-D). A and B share the color bar in the top row; C and D share the color bar in the bottom row. Detailed counts are available in Supplementary Tables \ref{tab:countries-50}-\ref{tab:countries-125}}
    \label{fig:country_distributions}
\end{figure}

In this paper, we use GLiNER to showcase performance enhancements resulting from our proposed methodology for grounding and diversifying synthetic data. Our approach leverages a voluminous collection of news articles across 125 countries (Figure \ref{fig:country_distributions}A) and in 12 languages, ensuring a rich tapestry of global perspectives and narratives. This diverse base is further augmented through processes like topic diversification, translation, and summarization, thereby creating a synthetic dataset that not only supports robust model training but also significantly reduces the bias inherent in more homogenous datasets. We present a detailed description of the methodology and show that the resulting dataset improves performance on common NER benchmark datasets by up to 7.3\%. Further, we provide the open-sourced dataset, comprising 5,049 samples of summarized news articles covering 73 unique topics and 54 unique entity types, as well as fine-tuned models available under Apache 2.0 license.
Initially applied to the domain of NER, our methodology demonstrates a scalable strategy for enhancing data representativeness in numerous AI disciplines.

\section{Related work}

\citet{veselovsky2023generating} investigated three strategies for improving faithfulness in synthetically generated data: grounding, filtering, and taxonomy-based generation. Applied to the task of identifying sarcasm, they concluded that grounding the prompt with a real example text resulted in the highest improvement. \citet{josifoski2023exploiting} created synthetic data for closed information extraction (cIE) using graph triplets and showed that cIE models trained on this curated and grounded dataset outperformed models trained on manually annotated data. \citet{eldan2023tinystories} enforced diversity in LLM-generated text by incorporating a grounding vocabulary and showed how training small language models (SMLs) on the resulting dataset improved creativity in the SLMs' output.

\section{Method}
\label{sec:method}

\begin{figure}[h!]
    \centering
    \includegraphics[width=0.95\textwidth]{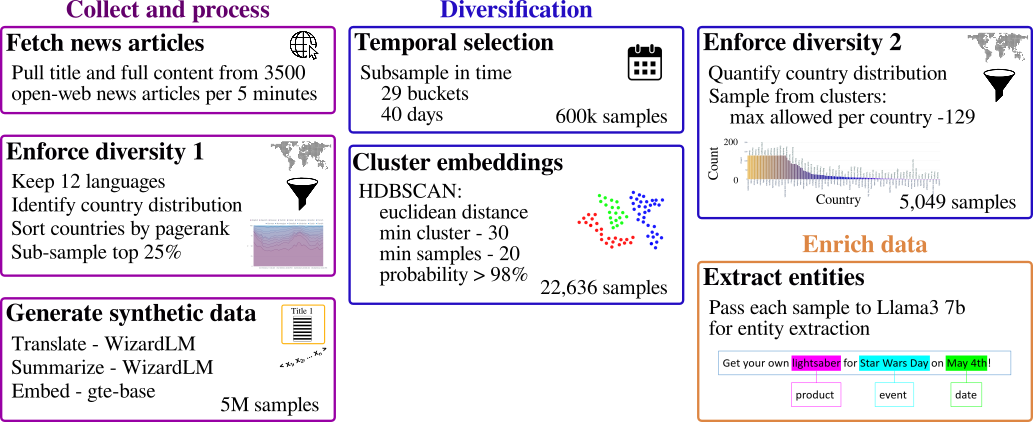}
    \caption{\textbf{Dataset engineering.} Procedure used for enforcing diversity and generating synthetic data for AskNews-NER-v0.}
    \label{fig:dataset_engineering}
\end{figure}

This section details the steps taken to diversify the news data underpinning the grounded synthetic data generation, the data generation itself, as well as the selection of data for the NER dataset. The full procedure is shown in Figure \ref{fig:dataset_engineering}.

\subsection{Synthetic data diversification}

We parsed close to 1 million open-web, non-paywalled, news articles per day from news outlets across 212 countries. Every five minutes, approximately 3,500 articles are passed through the following pipeline:
\begin{enumerate}
    \item Keep all articles corresponding to the following 12 languages: English, Spanish, Portuguese, German, Russian, French, Arabic, Italian, Ukrainian, Norwegian, Swedish, and Danish (this list is determined by the LLM used in Step 5 for translation and summarization).
    \item Identify the distribution of country of origin for the news outlets in the selected articles (e.g., Figure \ref{fig:country_distributions}A)
    \item Sort the articles by Majestic Million page rank, which preferences outlets with more backlinks.
    \item Sample a subset of articles, enforcing the country of origin distribution from Step 2.
    \item Translate, summarize, and classify each article using WizardLM 13B V1.2 \citep{xu2023wizardlm}.
    \item Embed the titles and summaries using the \texttt{thenlper/gte-base} embedding model \citep{li2023towards}.
    \item Store the embeddings in a vector database (we are using Qdrant \citep{qdrant}).
\end{enumerate}

The vector database is accessible via an API endpoint as detailed here: \href{https://docs.asknews.app/en/reference#get-/v1/news/search}{AskNews API documentation}.

\subsection{The AskNews-NER-v0 dataset}
\label{subsec:dataset}

\paragraph{Topic diversity}

\begin{figure}[h!]
    \centering
    \includegraphics[width=\textwidth]{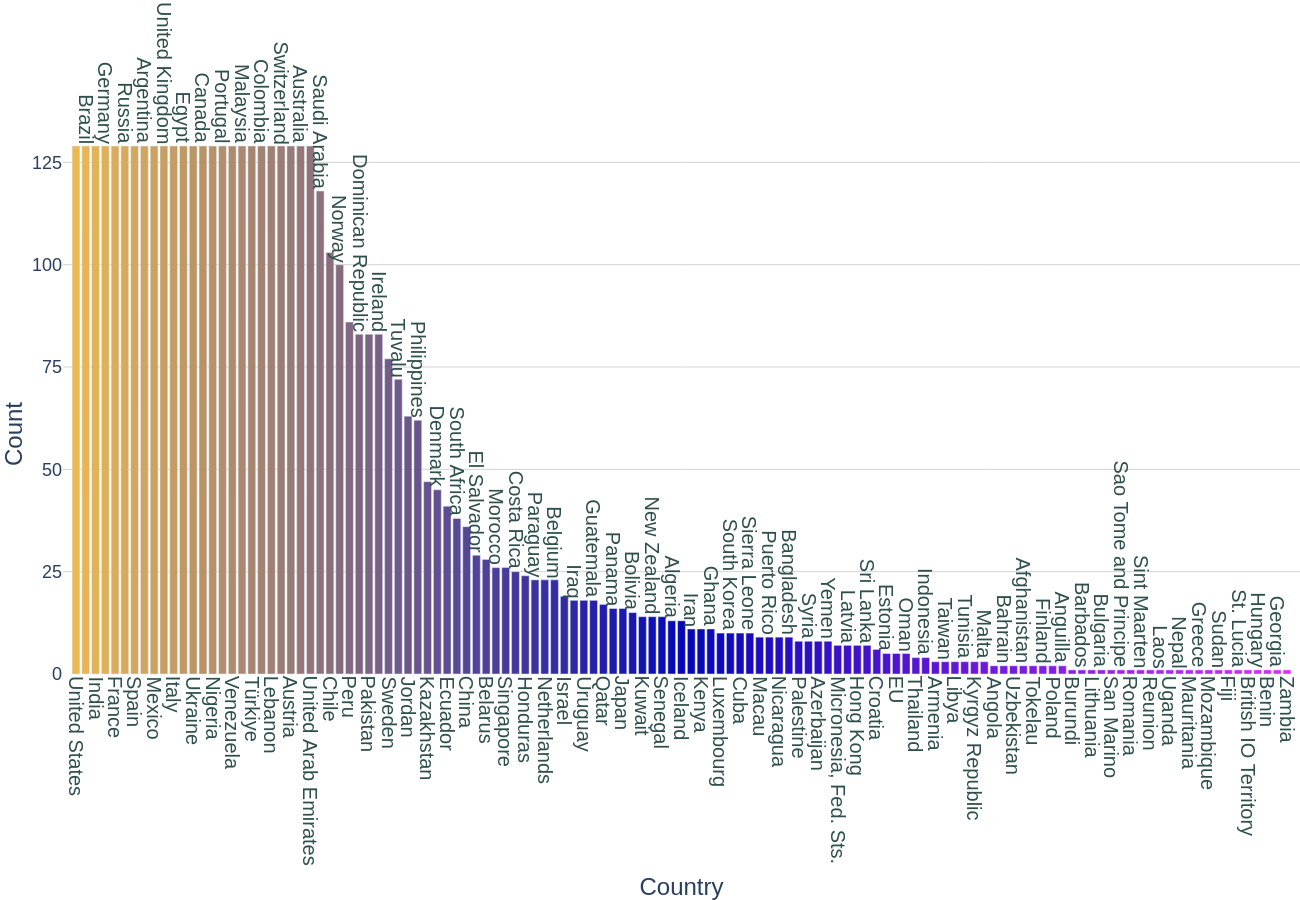}
    \caption{\textbf{Country representation.} Country representation in the AskNews-NER-v0 dataset after enforcing country distribution, as per step 6 of the topic diversification pipeline.}
    \label{fig:country-representation}
\end{figure}

Topic diversification was enforced by grouping the articles based on semantic similarity through clustering of their embeddings, then sampling from the clusters while enforcing country distributions. The topic diversification pipeline follows:

\begin{enumerate}
    \item Select articles indexed between Tuesday, February 20, 2024 6:10 PM  and Sunday, March 31, 2024 2:10 PM UTC. ~600k articles.
    \item Sub-sample 29 evenly spaced 4 hour buckets of embedded articles
    \item Cluster each bucket of embeddings after L2-normalizarion using scikit-learn's \citep{scikit-learn} HDBSCAN algorithm using the Euclidean distance metric, a minimum cluster size of 30, and a minimum sample size of 20 (595 total clusters were identified across all buckets, each cluster corresponding to semantically similar articles). 
    \item Filter embeddings for cluster probabilities > 98\% (the resulting clusters contained between 30 and 112 articles, totalling 22,636 articles).
    \item Identify the distribution of countries. 22,636 articles.
    \item Enforce country distribution by ensuring to sample low-representation countries and not over-sampling any high-representation country (Figure \ref{fig:country-representation}). 5,049 articles.
\end{enumerate}

\begin{figure}[h!]
    \centering
    \includegraphics[width=0.65\textwidth]{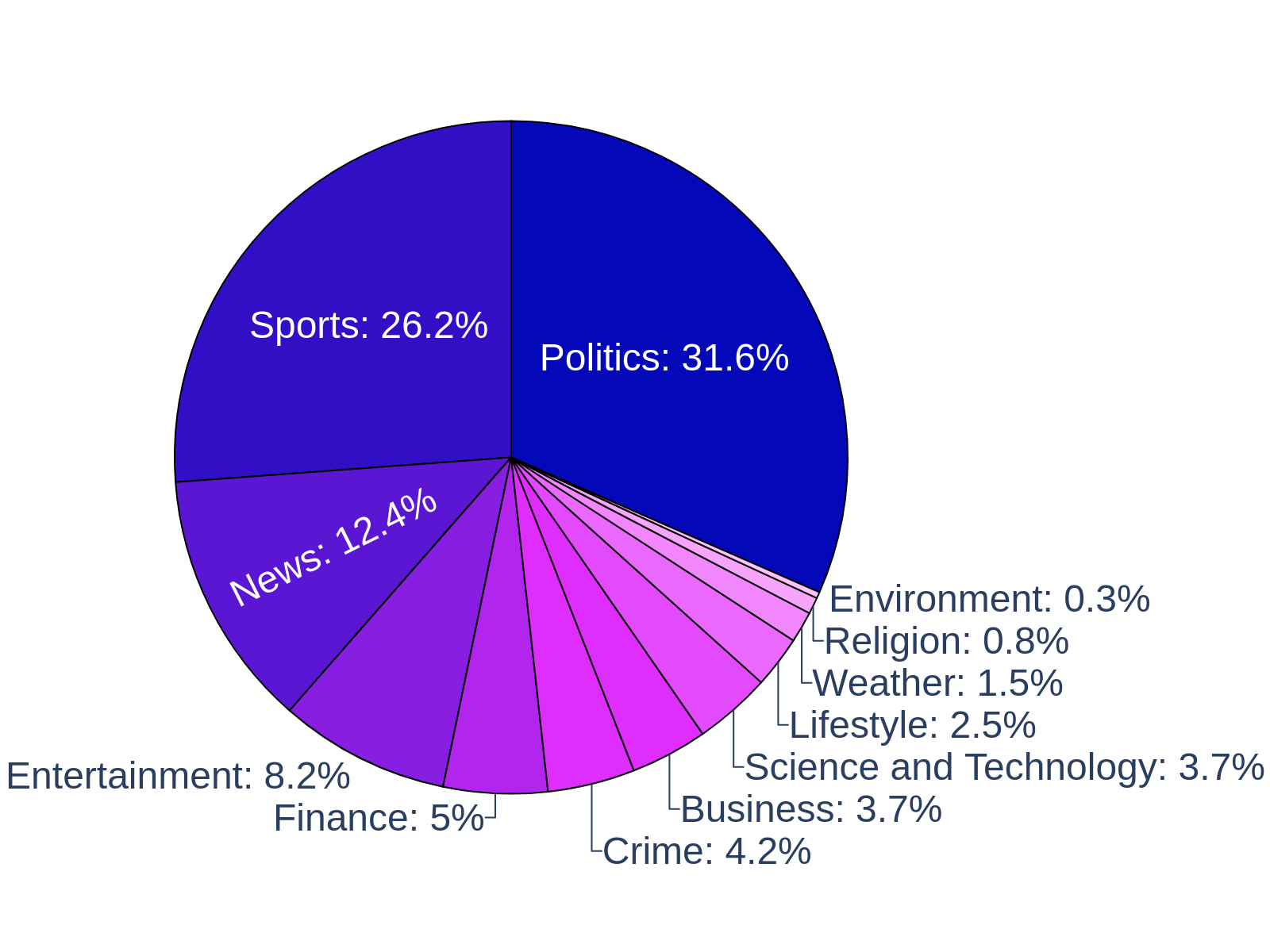}
    \caption{\textbf{Topics distribution.} Topics summarized by GPT4. The full set of topics are available in Supplementary Figure \ref{sup-fig:topics}.}
    \label{fig:topics_grouped}
\end{figure}

The procedure outlined above yielded 5,049 articles covering 73 unique topics. The top ten most frequent topics were (in descending order) Politics, Sports, News, Finance, Crime, Entertainment, Business, Technology, Weather, and Science (see Supplementary Figure \ref{sup-fig:topics} for details). For an easier overview, the 73 topics were grouped using GPT4 (Figure \ref{fig:topics_grouped}).

\paragraph{Entity labeling}

We used \texttt{meta-llama/Meta-Llama-3-70B-Instruct} \citep{llama3modelcard} for labeling entities in the news articles. The instructions provided to the LLM are shown in Figure \ref{box:prompt}. We restricted the requested entity types to 54 choices as shown in Figure \ref{box:types}.

\begin{center}
\captionof{figure}{\textbf{Entity extraction prompt.} We gave the following instructions to Llama-3-70B-Instruct regarding identification and labeling of entities. We also provided two example input texts and expected outputs, as shown in Supplementary Figure \ref{sup-box:prompt}.}
\label{box:prompt}
\noindent\fcolorbox{black}{bluebg}{%
\begin{minipage}{\dimexpr0.95\textwidth-2\fboxrule-2\fboxsep\relax}
\paragraph{Prompt}A chat between a User and an artificial intelligence Assistant that is an expert at identifying and extracting named entities from text. The Assistant's task is to analyze a given text and extract all entities and identify their entity types.\\

Instructions for the Assistant:\\
1. **read the text** The Assistant reads the text provided by the User.\\
2. **identify entities** As the Assistant reads the text, the Assistant identifies entities in the text in the order that they exist in the text.\\
3. **determine entity types** For each Found Entity, the Assistant determines an entity type that best describes the Found Entity from the following list of entity types:
\{entity\_types\}\\
4. **verbatim extraction** The Assistant extracts the Found Entity exactly as found in the text. The Assistant does not invent entities that are not present in the text.\\

The output should be a JSON dictionary according to the following schema:
\{\{'properties': \{\{'entities': \{\{'items': \{\{'items': \{\{'type': 'string'\}\}, 'maxItems': 2, 'minItems': 2, 'type': 'array'\}\}, 'title': 'Entities', 'type': 'array'\}\}\}\}, 'required': ['entities'], 'title': 'ENTITY\_MODEL', 'type': 'object'\}\}\\

Example:
Text Input: "In 2021, Apple released the iPhone 13 at 75\% battery power."
Expected Output: \{\{ "entities" : [ [ "2021" , "Date" ] , [ "Apple" , "Organization" ] , [ "iPhone 13" , "Product" ] , [ "75\%" , "Percentage" ] ] \}\}\\

Given the following text, please extract all entities and their entity types:\\
<doc>
\{input\}
</doc>
\end{minipage}}
\end{center}

\begin{center}
\captionof{figure}{\textbf{Requested entity types.} We specified 54 entity types that we requested Llama-3-70B-Instruct to identify and label in the news articles. The order was shuffled for each article to reduce the risk of influencing the labeling.}
\label{box:types}
\noindent\fcolorbox{black}{bluebg}{%
\begin{minipage}{\dimexpr0.95\textwidth-2\fboxrule-2\fboxsep\relax}
\paragraph{Entity types}Person, Politician, Actor, Athlete, Artist, Writer, Character, Musician, Scientist, Director, Title, Organization, Political Party, Election, Facility, Location, Country, City, Nationality, Language, Year, Date, Time, Event, Award, Song, Movie, Book, Media, Band, Album, TV Show, Game, Software, Brand, Money, Price, Law, Quantity, Number, Percentage, Sports, Politics, Weapon, Product, Food, Material, Transportation, Vehicle, Religion, Technology, Space, Medicine, Science. 
\end{minipage}}
\end{center}

\begin{figure}[h!]
    \centering
    \includegraphics[width=0.95\textwidth]{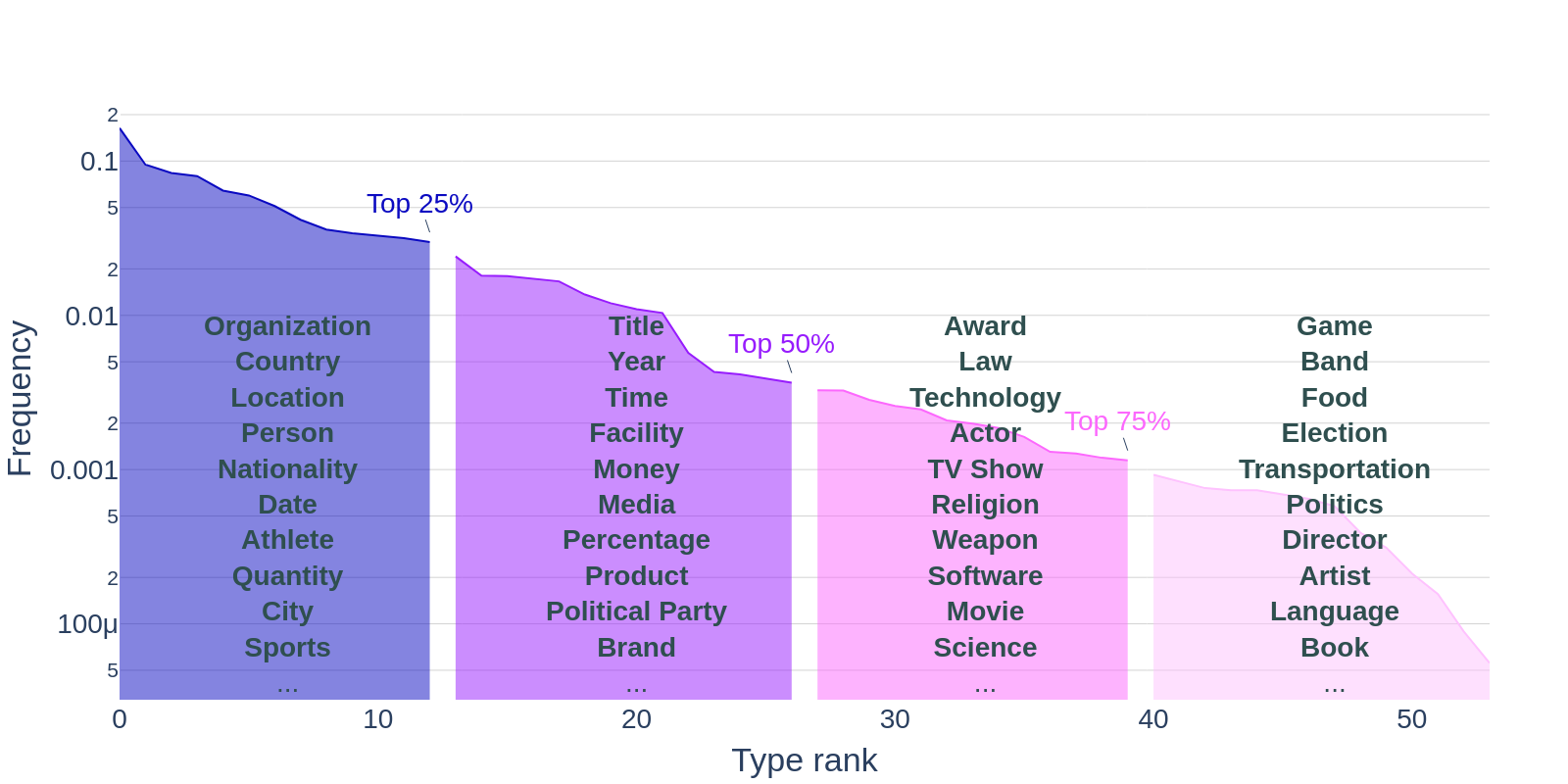}
    \caption{\textbf{Entity type frequency for requested labels.}}
    \label{fig:types-limited}
\end{figure}

Despite the instructions to limit the entity types, the labeled articles sported 182 unique entity types (Supplementary Figure \ref{sup-fig:types-unlimited}). Entity types that were not requested were subsequently removed from the AskNews-NER-v0 dataset, resulting in the distribution seen in Figure \ref{fig:types-limited}.

\paragraph{Dataset splits}
The dataset was split temporally such that data from the four latest time windows (between Thursday, March 28, 2024 2:10 PM and Wednesday, March 27, 2024 8:10 AM) was selected as test data, amounting to 730 samples. The remaining data was stratified based on time window and cluster. 730 samples, to be used as the validation set, were selected such that each time window and cluster were sampled at least once. The remaining data, amounting to 3,589 samples, was assigned as training data. Country distributions for the splits are shown in Figure \ref{fig:country_distributions}B-D. Detailed counts are available in Supplementary Tables \ref{tab:countries-50}-\ref{tab:countries-125}. 

\subsection{GLiNER fine-tuning}
\label{subsec:fine-tuning}

\paragraph{Training hyperparameters}
We fine-tuned GLiNER-Small-v2.1 (50M parameters), GLiNER-Medium-v2.1 (90M parameters), and GLiNER-Large-v2.1 (0.3B parameters) on the AskNews-NER-v0 dataset. All three models were trained on the NuNER dataset \citep{bogdanov2024nuner} and are available under the Apache 2.0 license. 

Training was done using the AdamW optimizer. The base learning rate for both pretrained layers and non-pretrained layers was set to 1e-6 and adjusted using \texttt{ReduceLROnPlateau} with factor 0.5 and patience 5. The models were trained for 25 epochs, with a batch size of 5, for 1,000 steps.
Evaluation using the validation data was done with a batch size of 12 and a span probability threshold of 0.5 every epoch, using the top 30 most common entity types present in the news dataset (see Supplementary Table W for details). Both shuffling and random drop of entity types were enabled for regularization purposes \citep{sainz2023gollie}. The best model was chosen based on the highest micro-F1 score on the evaluation data. 

\paragraph{Compute}
Training was done using a NVIDIA RTX A4500 with 20GB RAM and took $\sim$1h 15min, $\sim$1h 48min, and $\sim$4h 3min for the small, medium, and large GLiNER models, respectively.

\subsection{Evaluation}

\paragraph{Datasets}
We evaluated our fine-tuned models in a zero-shot context on common NER benchmarks, as were presented in \citep{zaratiana2023gliner}. These benchmark datasets include the Out-of-Domain (OOD) NER Benchmark (Table \ref{tab:eval-ood}), which contains seven NER datasets from the CrossNER \citep{liu2020crossner} and MIT \citep{liu2013asgard, liu2013query} benchmark collections, as well as 18 NER datasets (Table \ref{tab:eval-20}) that are commonly used to train NER models. 

\paragraph{Models}
We compared the zero-shot performance on the aforementioned datasets between the base GLiNER models, i.e., Small-v1, Medium-v1, Large-v1 (trained on the Pile-NER dataset \citep{zhou2023universalner} and available under CC BY NC 4.0 license), and Small-v2.1, Medium-v2.1, and Large-v2.1 (trained on the NuNER dataset \citep{bogdanov2024nuner} and available under Apache 2.0 license).

\paragraph{Metric}
Evaluation was done using micro-F1 score, computed based on the exact match between predicted and ground-truth entity types.

\paragraph{Compute}
Evaluation was performed using a NVIDIA GeForce RTX 3090 with 32GB of RAM and took $\sim$6min, $\sim$8min, and $\sim$15min for the small, medium, and large GLiNER models, respectively.

\section{Results}
\label{sec:results}

The quality of our AskNews-NER-v0 dataset was assessed through micro-F1 scores on 18 common NER benchmark datasets after fine-tuning of the commercially permissible GLiNER models, i.e., GLiNER-Small-v2.1, GLiNER-Medium-v2.1, and GLiNER-Large-v2.1. 

Tables \ref{tab:eval-20} and \ref{tab:eval-ood} show that our fine-tuned models improve the performance of the base models by 4.6\%, on average, and outperforms other LLM-based models. The fine-tuned small model showed increased scores of 2.5 points (5.5\%) on the 18 NER datasets and 0.9 points (1.7\%) on the OOD benchmarks. The scores for the fine-tuned medium model also showed improvements with an additional 3.3 points (7.2\%) on the 18 NER datasets and 2.1 points (3.6\%) on the OOD benchmarks. Similarly, the fine-tuning of the large model improved the score by and 3.3 points (7.3\%) on the 18 NER datasets and 1 point (1.7\%) on the OOD benchmarks. 

Note that the performance of the GLiNER models detailed in \citet{zaratiana2023gliner} does not align perfectly with the scores presented here in Tables \ref{tab:eval-20} and \ref{tab:eval-ood} as the GLiNER models were refactored before publication to Huggingface (see discussion in GitHub Issue here: \href{https://github.com/urchade/GLiNER/issues/54}{ Reproduction questions \#54}). As our models were fine-tuned on the Huggingface versions, we evaluate those models rather than present the previous scores. A comparison between the scores published in \citet{zaratiana2023gliner} and in this paper can be found in Supplementary Tables \ref{sup-tab:hf-ood} and \ref{sup-tab:hf-20}.

\begin{table}[h!]
\caption{\textbf{Zero-shot performance on 18 NER datasets.} Performance comparison between NuNERZero-span, GLiNER base models ("v2.1"), and fine-tuned models ("News") on common NER benchmarks as well as on the test split of the AskNews-NER-v0 dataset. \textbf{Bold} indicates the highest score per dataset. \textcolor{green}{Green} and \textcolor{red}{red} values indicate increase or decrease in score for the fine-tuned model compared to the base model. Comparison also to GLiNER-v1 models are found in Supplementary Table \ref{sup-tab:eval-20-all}. Commit hashes for all models are found in Supplementary Table \ref{sup-tab:commits}.}
\label{tab:eval-20}
\resizebox{\textwidth}{!}{%
\centering
\begin{tabular}{c|c|cc|cc|cc}
\toprule
Dataset & NuNerZero-span & S-v2.1 & S-News & M-v2.1 & M-News & L-v2.1 & L-News \\ \midrule
ACE 2005 & 23.6 & 32.8 & 23.2 \tiny{\textcolor{red}{-9.6}} & 31.3 & 23.3 \tiny{\textcolor{red}{-8.0}} & \textbf{33.3} & 28.9 \tiny{\textcolor{red}{-4.4}} \\ 
AnatEM & 29.2 & 35.1 & 36.5 \tiny{\textcolor{green}{+1.4}} & 29.2 & \textbf{36.8} \tiny{\textcolor{green}{+7.6}} & 24.7 & 34.8 \tiny{\textcolor{green}{+10.1}} \\ 
Broad Tweet Corpus & 60.2 & 63.0 & \textbf{69.2} \tiny{\textcolor{green}{+6.2}} & 63.9 & 67.7 \tiny{\textcolor{green}{+3.8}} & 59.9 & \textbf{69.2} \tiny{\textcolor{green}{+9.3}} \\ 
CoNLL 2003 & \textbf{63.6} & 59.8 & 60.9 \tiny{\textcolor{green}{+1.1}} & 61.5 & 63.0 \tiny{\textcolor{green}{+1.5}} & 59.5 & 56.4 \tiny{\textcolor{red}{-3.1}} \\ 
FabNER & \textbf{24.0} & 17.9 & 18.9 \tiny{\textcolor{green}{+1.0}} & 17.0 & 20.6 \tiny{\textcolor{green}{+3.6}} & 19.2 & 22.5 \tiny{\textcolor{green}{+3.3}} \\ 
FindVehicle & 43.7 & 38.0 & 43.1 \tiny{\textcolor{green}{+5.1}} & 36.0 & 38.7 \tiny{\textcolor{green}{+2.7}} & 51.9 & \textbf{52.9} \tiny{\textcolor{green}{+1.0}} \\ 
GENIA\_NER & 55.0 & 47.7 & 47.2 \tiny{\textcolor{red}{-0.5}} & \textbf{55.4} & 54.0 \tiny{\textcolor{red}{-1.4}} & 55.3 & 55.0 \tiny{\textcolor{red}{-0.3}} \\ 
HarveyNER & 24.9 & 18.3 & 23.1 \tiny{\textcolor{green}{+4.8}} & 23.2 & \textbf{26.7} \tiny{\textcolor{green}{+3.5}} & 18.7 & 15.8 \tiny{\textcolor{red}{-2.9}} \\ 
MultiNERD & 63.9 & 57.3 & \textbf{68.3} \tiny{\textcolor{green}{+11.0}} & 55.3 & 67.1 \tiny{\textcolor{green}{+11.8}} & 48.7 & 64.0 \tiny{\textcolor{green}{+15.3}} \\ 
Ontonotes & 31.6 & 28.1 & \textbf{35.1} \tiny{\textcolor{green}{+7.0}} & 25.7 & 33.2 \tiny{\textcolor{green}{+7.5}} & 19.5 & 32.0 \tiny{\textcolor{green}{+12.5}} \\ 
PolyglotNER & 42.8 & 40.4 & 43.9 \tiny{\textcolor{green}{+3.5}} & 42.1 & \textbf{45.3} \tiny{\textcolor{green}{+3.2}} & 39.6 & 42.5 \tiny{\textcolor{green}{+2.9}} \\ 
TweetNER7 & \textbf{40.1} & 36.5 & 35.6 \tiny{\textcolor{red}{-0.9}} & 38.2 & 38.6 \tiny{\textcolor{green}{+0.4}} & 36.1 & 35.5 \tiny{\textcolor{red}{-0.6}} \\ 
WikiANN en & \textbf{58.1} & 55.3 & 53.6 \tiny{\textcolor{red}{-1.7}} & 55.5 & 53.0 \tiny{\textcolor{red}{-2.5}} & 55.5 & 52.5 \tiny{\textcolor{red}{-3.0}} \\ 
WikiNeural & 72.3 & 64.7 & 76.0 \tiny{\textcolor{green}{+11.3}} & 67.5 & \textbf{77.9} \tiny{\textcolor{green}{+10.4}} & 62.9 & 70.3 \tiny{\textcolor{green}{+7.4}} \\ 
bc2gm & 52.7 & 50.4 & 49.4 \tiny{\textcolor{red}{-1.0}} & 54.0 & \textbf{54.6} \tiny{\textcolor{green}{+0.6}} & 45.2 & 46.9 \tiny{\textcolor{green}{+1.7}} \\ 
bc4chemd & 50.8 & 49.0 & 49.4 \tiny{\textcolor{green}{+0.4}} & 47.3 & 51.1 \tiny{\textcolor{green}{+3.8}} & 53.1 & \textbf{55.1} \tiny{\textcolor{green}{+2.0}} \\ 
bc5cdr & 69.7 & 66.1 & 70.9 \tiny{\textcolor{green}{+4.8}} & 67.0 & \textbf{72.5} \tiny{\textcolor{green}{+5.5}} & 68.4 & 71.2 \tiny{\textcolor{green}{+2.8}} \\ 
ncbi & 61.4 & 56.6 & 57.9 \tiny{\textcolor{green}{+1.3}} & 60.3 & 64.5 \tiny{\textcolor{green}{+4.2}} & 59.7 & \textbf{65.5} \tiny{\textcolor{green}{+5.8}} \\ 
\midrule
Average & 48.2 & 45.4 & 47.9 \tiny{\textcolor{green}{+2.5}} & 46.1 & \textbf{49.4} \tiny{\textcolor{green}{+3.3}} & 45.1 & 48.4 \tiny{\textcolor{green}{+3.3}} \\ 
\midrule
\midrule
 AskNews-NER-v0 &  61.1 &  54.4 &  71.8 \tiny{\textcolor{green}{+17.4}} &  53.3 &  71.9 \tiny{\textcolor{green}{+18.6}} &  48.6 &  \textbf{72.9} \tiny{\textcolor{green}{+24.3}} \\ 
\bottomrule
\end{tabular}}
\end{table}

\begin{table}[h!]
\caption{\textbf{Zero-Shot Scores on Out-of-Domain NER Benchmarks.} Performance for LLM-based models and comparison between base models ("v2.1") and fine-tuned models ("News"). \textbf{Bold} indicates the highest score per dataset. \textcolor{green}{Green} and \textcolor{red}{red} values indicate increase or decrease in score for the fine-tuned model compared to the base model. Results for ChatGPT and UniNER are from \citep{zhou2023universalner}; InstructUIE is from \citep{wang2023instructuie}; GoLLIE is from \citep{sainz2023gollie}. Huggingface \includegraphics[scale=0.06]{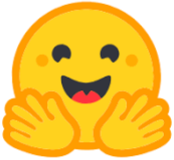} commit hashes for the NuNER model and GLiNER models are found in Supplementary Table \ref{sup-tab:commits}.}
\label{tab:eval-ood}
\resizebox{\textwidth}{!}{%
\centering
\begin{tabular}{c|lllllll|l}
\toprule
Model & AI & Literature & Music & Politics & Science & Movie & Restaurant & Average \\
\midrule
ChatGPT & 52.4 & 39.8 & 66.6 & 68.5 & 67.0 & 5.3 & 32.8 & 47.5 \\ 
InstructUIE & 49.0 & 47.2 & 53.2 & 48.1 & 49.2 & \textbf{63.0} & 21.0 & 47.2 \\ 
UniNER-7B & 53.6 & 59.3 & 67.0 & 60.9 & 61.1 & 42.4 & 31.7 & 53.7 \\ 
UniNER-13B & 54.2 & 60.9 & 64.5 & 61.4 & 63.5 & 48.7 & 36.2 & 55.6 \\ 
GoLLIE & 59.1 & 62.7 & 67.8 & 57.2 & 55.5 & \textbf{63.0} & 43.4 & 58.0 \\ 
NuNerZero-span & \textbf{60.1} & \textbf{64.9} & 69.9 & \textbf{72.7} & 66.4 & 55.0 & 41.1 & \textbf{61.5} \\ 
\midrule
\midrule
S-v1 & 50.7 & 60.0 & 60.9 & 61.5 & 55.6 & 46.9 & 33.3 & 52.7 \\ 
S-v2.1 & 52.6 & \textbf{64.9} & 66.7 & 64.3 & 64.3 & 47.2 & 20.8 & 54.4 \\ 
S-News & 57.0 \tiny{\textcolor{green}{+4.4}} & 63.5 \tiny{\textcolor{red}{-1.4}} & 64.7 \tiny{\textcolor{red}{-2.0}} & 65.6 \tiny{\textcolor{green}{+1.3}} & 63.9 \tiny{\textcolor{red}{-0.4}} & 46.9 \tiny{\textcolor{red}{-0.3}} & 25.7 \tiny{\textcolor{green}{+4.9}} & 55.3 \tiny{\textcolor{green}{+0.9}} \\ 
\midrule
M-v1 & 51.8 & 59.7 & 69.4 & 68.6 & 58.1 & 42.9 & 37.3 & 55.4 \\ 
M-v2.1 & 52.0 & 62.6 & 68.9 & 65.7 & 65.2 & 46.5 & 30.9 & 56.0 \\ 
M-News & 57.1 \tiny{\textcolor{green}{+5.1}} & 64.3 \tiny{\textcolor{green}{+1.7}} & \textbf{70.3} \tiny{\textcolor{green}{+1.4}} & 67.1 \tiny{\textcolor{green}{+1.4}} & \textbf{67.9} \tiny{\textcolor{green}{+2.7}} & 45.2 \tiny{\textcolor{red}{-1.3}} & 35.1 \tiny{\textcolor{green}{+4.2}} & 58.1 \tiny{\textcolor{green}{+2.1}} \\ 
\midrule
L-v1 & 57.2 & 64.4 & 69.6 & 72.6 & 62.6 & 57.2 & 42.9 & 60.9 \\ 
L-v2.1 & 53.4 & 57.9 & 67.1 & 64.9 & 62.0 & 51.3 & \textbf{46.0} & 57.5 \\ 
L-News & 57.0 \tiny{\textcolor{green}{+3.6}} & 60.4 \tiny{\textcolor{green}{+2.5}} & 65.2 \tiny{\textcolor{red}{-1.9}} & 62.9 \tiny{\textcolor{red}{-2.0}} & 65.9 \tiny{\textcolor{green}{+3.9}} & 56.6 \tiny{\textcolor{green}{+5.3}} & 41.8 \tiny{\textcolor{red}{-4.2}} & 58.5 \tiny{\textcolor{green}{+1.0}} \\ 
\bottomrule
\end{tabular}}
\end{table}

\section{Discussion}
\label{sec:discussion}

In this paper, we presented a methodology for grounding synthetic data in real-world diversity by using news texts originating from global sources that cover numerous topics. Using this methodology, we curate a diverse dataset for Named Entity Recognition (NER) and demonstrated that training NER models on this grounded, synthetic dataset significantly enhances model performance by enabling the models to recognize and generalize across a broader spectrum of linguistic and cultural nuances. 

The diversity of data sources is pivotal in developing robust AI systems. By integrating news texts from a wide geographical and linguistic spread, we captured diverse contexts that are often underrepresented. 
This approach addresses critical gaps in common dataset that tend to be biased towards more readily available data sources. 
As a result, models trained on diversified and grounded data are better equipped to handle real-world applications across different languages and regions, reducing the cultural and linguistic biases prevalent in many current AI systems.
Our methodology incorporated topic diversification, translation, and summarization to enrich the synthetic dataset. 
These strategies collectively improve the quality of the generated dataset and can be implemented to create data for real-world tasks beyond NER, such as text classification and sentiment analysis. In fact, while the dataset described in this paper is curated for diversity, the methodology can easily be extended to domain-specific text corpora as the framework used incorporates a vector database containing both sparse and dense text embeddings, hence allowing for key-word as well as natural language queries to obtain articles within a specific domain. 

The grounding of synthetic datasets in diversified real-world data introduces significant advancements with both substantial benefits and noteworthy challenges. On the positive side, these datasets enhance the inclusivity and fairness of AI systems. By reflecting a broader array of cultural and linguistic backgrounds, they help create models that perform equitably across diverse populations, reducing biases that traditionally affect AI implementations. This contributes to more reliable and universally applicable AI solutions, fostering greater trust and broader adoption across various regions and communities. However, the potential for perpetuating existing biases in source materials does exist, particularly if certain regions or groups are underrepresented in global news coverage. Additionally, the complexities of translating and summarizing content from various languages might introduce inaccuracies. Yet, these challenges highlight the importance of refining data synthesis techniques and establishing rigorous oversight processes. By continually improving these methodologies, the field can leverage the full potential of diversified data to develop AI systems that are not only more effective but also truly equitable and beneficial for a global audience. 

\subsection{Limitations}
\label{subsec:limitations}

\paragraph{Language limitations}
Our methodology is currently limited to 12 languages due to the choice of LLM for translation and summarization.It is possible to extend the method to cover more languages by, e.g., employing multiple LLMs, each with different language coverage, and adaptively choosing which one to use based on the language of the news article.  Furthermore, the accuracy of translations and the risk of losing nuances in summarization are concerns that could affect the fidelity of the synthetic data. 

\paragraph{Model bias}
Although the goal of the dataset is to reduce bias, and improve diversity, it is still biased to western languages and countries. This limitation originates from the abilities of WizardLM 13B v1.2 for the translation and summary generations. Further, any bias originating in WizardLM 13B v1.2 training data will also be present in this dataset, since WizardLM 13B v1.2 was used to summarize the open-web articles. Further, any biases present in Llama-3-70B-Instruct will be present in the present dataset since Llama-3-70B-Instruct was used to extract entities from the summaries

\paragraph{Model availability}
All models used in the creation of the dataset, namely WizardLM 13B V1.2 and Llama-3-70B-Instruct are open-sourced and available both via API endpoints and for running locally. Similarly, the GLiNER models used for evaluating our methodology are open-sourced and available on Huggingface. Whilst the GLiNER models are compact and can be run on most hardware, the Llama models require more compute or paying for API calls. 

\section{Conclusion}

Our research presents a compelling blueprint for generation of diversified synthetic data aimed at training AI models. By grounding synthetic data in real-world diversity, we enhance the generalizability and fairness of AI systems. This study not only advances our understanding of effective dataset synthesis but also underscores the critical need for diversity in training AI to tackle real-world challenges across various domains.

\paragraph{Data, model, and code availability}
The AskNews-NER-v0 dataset and fine-tuned NER models are available on Huggingface at \href{https://huggingface.co/datasets/EmergentMethods/AskNews-NER-v0}{AskNews-NER-v0} and \href{https://huggingface.co/EmergentMethods/gliner_small_news-v2.1}{GLiNER-Small-News}, \href{https://huggingface.co/EmergentMethods/gliner_medium_news-v2.1}{GLiNER-Medium-News}, and \href{https://huggingface.co/EmergentMethods/gliner_large_news-v2.1}{GLiNER-Large-News}, respectively, under Apache 2.0 license. The code used for the analysis in this paper is available on GitHub at *not yet available*.

\newpage

\bibliographystyle{unsrtnat}
\bibliography{references}

\newpage
\appendix

\renewcommand{\figurename}{Supplementary Figure}
\setcounter{figure}{0}
\renewcommand{\tablename}{Supplementary Table}
\setcounter{table}{0}

\section{Appendix}

\subsection{Entity labeling}

We used \texttt{meta-llama/Meta-Llama-3-70B-Instruct} \citep{llama3modelcard} for labeling entities in the news articles. The instructions provided to the LLM are shown in Figure \ref{box:prompt}. We restricted the requested entity types to 54 choices as shown in Figure \ref{box:types}. In addition to the prompt instructions, we provided the following examples (Supplementary Figure \ref{sup-box:prompt}) to guide the LLM.

\begin{center}
\captionof{figure}{\textbf{Entity extraction prompt examples.} We provided the following two example input texts and expected outputs, in addition to the prompt in Figure \ref{box:prompt}.}
\label{sup-box:prompt}
\noindent\fcolorbox{black}{bluebg}{%
\begin{minipage}{\dimexpr0.95\textwidth-2\fboxrule-2\fboxsep\relax}
\paragraph{Prompt examples}
Q: Given the following text, please extract all named entities:
<doc>
French Prime Minister Gabriel Attal, 61, announced at the ongoing United Nations (UN) climate summit that France will be investing £500 million into the French renewable energy sector, specifically, a Parisian hydroelectric dam. The news comes after the conservative French government's recent election victory and new laws being added to the French constitution.
</doc>\\

A: \{ "entities" : [ [ "French" , "Nationality" ] , [ "Prime Minister" , "Title" ] , [ "French Prime Minister" , "Title" ] , [ "French Prime Minister Gabriel Attal" , "Politician" ] , [ "Gabriel Attal" , "Person" ] , [ "61" , "Number" ] , [ "United Nations" , "Organization" ] , [ "UN" , "Organization" ] , [ "United Nations (UN) climate summit" , "Event" ] , [ "climate summit" , "Event" ] , [ "France" , "Country" ] , [ "£500 million" , "Money" ] , [ "French" , "Nationality" ] , [ "Parisian" , "Nationality" ] , [ "hydroelectric dam" , "Facility" ] , [ "French" , "Nationality" ] , [ "French government" , "Organization" ] , [ "election victory" , "Event" ] , [ "French" , "Nationality" ] , [ "French constitution" , "Law" ] , [ "constitution" , "Law" ] ] \}\\

Q: Given the following text, please extract all named entities:
<doc>
Daniel is 5-foot-2 and 181 pounds, while his brother, John, is 6-foot-1 and 200 pounds. They both play for the New York Giants in the NFL. Both scored 3 touchdowns, 10\% of the match's total, in the last game against the Dallas Cowboys, ending with a score of 21-17 after 45 minutes. Their average scores for the season are 12.5 points per game, after 4,000+ games. The first 5 games had -56 odds for the home team.
</doc>\\

A: \{ "entities" : [ [ "Daniel" , "Athlete" ] , [ "5-foot-2" , "Quantity" ] , [ "181 pounds" , "Quantity" ] , [ "John" , "Athlete" ] , [ "6-foot-1" , "Quantity" ] , [ "200 pounds" , "Quantity" ] , [ "New York" , "City" ] , [ "New York Giants" , "Organization" ] , [ "New York Giants" , "Sports" ] , [ "NFL" , "Organization" ] , [ "NFL" , "Sports" ] , [ "3 touchdowns" , "Quantity" ] , [ "10\%" , "Percentage" ] , [ "Dallas" , "City" ] , [ "Dallas Cowboys" , "Organization" ] , [ "Dallas Cowboys" , "Sports" ] , [ "21-17" , "Quantity" ] , [ "45 minutes" , "Time" ] , [ "12.5 points" , "Quantity" ] , [ "4,000+" , "Quantity" ] , [ "5" , "Quantity" ] , [ "-56" , "Number" ] ] \}
\end{minipage}}
\end{center}

\clearpage
\subsection{Dataset information}

\begin{figure}[h!]
    \centering
    \includegraphics[width=0.95\textwidth]{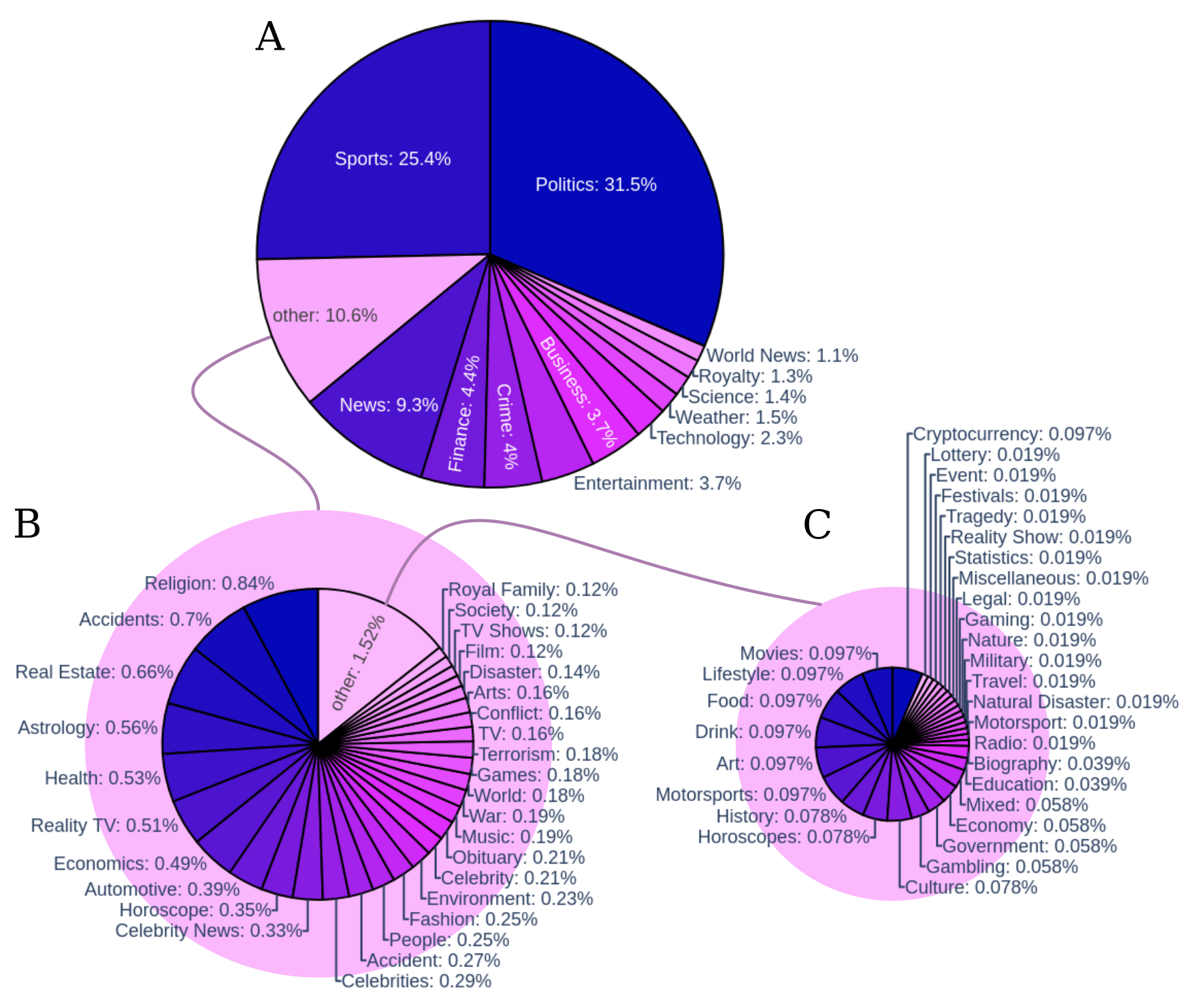}
    \caption{\textbf{Topics distribution.} Percentage coverage of the 73 unique topics assigned to the collection of articles used for the AskNews-NER-v0 dataset. For legibility, the topics have been split up into three separate charts where A shows topics with $\geq$1\% coverage and "other" corresponding to the remaining topics, B shows topics with $<$1\% but $\geq$0.1\% coverage and "other" corresponding to the remaining topics, and C shows topics with $<$0.1\% coverage.}
    \label{sup-fig:topics}
\end{figure}

Following topic diversification, as detailed in Section \ref{subsec:dataset}, the AskNews-NER-v0 dataset comprised 5,049 articles covering 73 unique topics (Supplementary Figure \ref{sup-fig:topics}) and originating from 125 different countries. The distribution of countries in the full dataset, as well as in the train, test, and validation splits, are shown in Supplementary Tables \ref{tab:countries-50}-\ref{tab:countries-125}. \texttt{meta-llama/Meta-Llama-3-70B-Instruct} \citep{llama3modelcard} was used for labeling entities in these news articles, resulting in the entity types shown in Supplementary Figure \ref{sup-fig:types-unlimited}. 

\begin{table}[h!]
\caption{\textbf{Country distribution.} Article frequency for the top 50 countries represented in the AskNews-NER-v0 dataset.}
\label{tab:countries-50}
\centering
\begin{tabular}{c|cccc}
\toprule
Country & Full dataset & Train split & Validation split & Test split \\ \midrule
United States & 129 & 86 & 22 & 21 \\ 
Canada & 129 & 100 & 20 & 9 \\ 
United Arab Emirates & 129 & 92 & 12 & 25 \\ 
Australia & 129 & 102 & 14 & 13 \\ 
Austria & 129 & 91 & 17 & 21 \\ 
Switzerland & 129 & 87 & 21 & 21 \\ 
Lebanon & 129 & 91 & 17 & 21 \\ 
Colombia & 129 & 91 & 16 & 22 \\ 
Türkiye & 129 & 92 & 13 & 24 \\ 
Malaysia & 129 & 87 & 18 & 24 \\ 
Brazil & 129 & 89 & 26 & 14 \\ 
Portugal & 129 & 103 & 17 & 9 \\ 
Nigeria & 129 & 101 & 21 & 7 \\ 
Venezuela & 129 & 87 & 24 & 18 \\ 
Ukraine & 129 & 91 & 13 & 25 \\ 
Spain & 129 & 93 & 22 & 14 \\ 
India & 129 & 79 & 24 & 26 \\ 
Egypt & 129 & 97 & 13 & 19 \\ 
France & 129 & 90 & 24 & 15 \\ 
Russia & 129 & 83 & 24 & 22 \\ 
Germany & 129 & 93 & 15 & 21 \\ 
Argentina & 129 & 87 & 24 & 18 \\ 
Mexico & 129 & 88 & 23 & 18 \\ 
United Kingdom & 129 & 93 & 25 & 11 \\ 
Italy & 129 & 99 & 16 & 14 \\ 
Saudi Arabia & 118 & 91 & 13 & 14 \\ 
Chile & 103 & 77 & 14 & 12 \\ 
Norway & 100 & 74 & 11 & 15 \\ 
Peru & 86 & 66 & 10 & 10 \\ 
Dominican Republic & 83 & 63 & 8 & 12 \\ 
Pakistan & 83 & 49 & 13 & 21 \\ 
Ireland & 83 & 59 & 15 & 9 \\ 
Sweden & 77 & 62 & 9 & 6 \\ 
Tuvalu & 72 & 43 & 16 & 13 \\ 
Jordan & 63 & 52 & 5 & 6 \\ 
Philippines & 62 & 27 & 7 & 28 \\ 
Kazakhstan & 47 & 30 & 4 & 13 \\ 
Denmark & 45 & 30 & 7 & 8 \\ 
Ecuador & 41 & 27 & 6 & 8 \\ 
South Africa & 38 & 30 & 4 & 4 \\ 
China & 36 & 26 & 7 & 3 \\ 
El Salvador & 29 & 19 & 5 & 5 \\ 
Belarus & 28 & 23 & 2 & 3 \\ 
Singapore & 26 & 14 & 5 & 7 \\ 
Morocco & 26 & 23 & 2 & 1 \\ 
Costa Rica & 25 & 19 & 4 & 2 \\ 
Honduras & 24 & 18 & 1 & 5 \\ 
Paraguay & 23 & 20 & 1 & 2 \\ 
Netherlands & 23 & 14 & 6 & 3 \\ 
Belgium & 23 & 20 & 3 & 0 \\ 
\end{tabular}
\end{table}

\begin{table}[h!]
\caption{\textbf{Country distribution.} Article frequency for the mid 51-100 countries represented in the AskNews-NER-v0 dataset. Continued from Supplementary Table \ref{tab:countries-50}.}
\label{tab:countries-100}
\centering
\begin{tabular}{c|cccc}
\toprule
Country & Full dataset & Train split & Validation split & Test split \\ \midrule
Israel & 19 & 10 & 7 & 2 \\ 
Guatemala & 18 & 12 & 1 & 5 \\ 
Iraq & 18 & 14 & 1 & 3 \\ 
Uruguay & 18 & 16 & 1 & 1 \\ 
Qatar & 17 & 12 & 5 & 0 \\ 
Panama & 16 & 12 & 3 & 1 \\ 
Japan & 16 & 9 & 2 & 5 \\ 
Bolivia & 15 & 10 & 2 & 3 \\ 
Kuwait & 14 & 8 & 4 & 2 \\ 
Senegal & 14 & 11 & 3 & 0 \\ 
New Zealand & 14 & 12 & 2 & 0 \\ 
Algeria & 13 & 9 & 1 & 3 \\ 
Iceland & 13 & 8 & 4 & 1 \\ 
Iran & 11 & 6 & 2 & 3 \\ 
Kenya & 11 & 7 & 1 & 3 \\ 
Ghana & 11 & 7 & 1 & 3 \\ 
Sierra Leone & 10 & 8 & 2 & 0 \\ 
Luxembourg & 10 & 8 & 0 & 2 \\ 
Cuba & 10 & 5 & 2 & 3 \\ 
South Korea & 10 & 8 & 1 & 1 \\ 
Macau & 9 & 7 & 1 & 1 \\ 
Puerto Rico & 9 & 6 & 2 & 1 \\ 
Nicaragua & 9 & 6 & 1 & 2 \\ 
Bangladesh & 9 & 6 & 3 & 0 \\ 
Yemen & 8 & 6 & 2 & 0 \\ 
Palestine & 8 & 5 & 2 & 1 \\ 
Azerbaijan & 8 & 6 & 1 & 1 \\ 
Syria & 8 & 6 & 0 & 2 \\ 
Sri Lanka & 7 & 4 & 3 & 0 \\ 
Hong Kong & 7 & 6 & 0 & 1 \\ 
Latvia & 7 & 2 & 2 & 3 \\ 
Micronesia, Fed. Sts. & 7 & 4 & 1 & 2 \\ 
Croatia & 6 & 6 & 0 & 0 \\ 
EU & 5 & 4 & 0 & 1 \\ 
Oman & 5 & 4 & 1 & 0 \\ 
Estonia & 5 & 3 & 1 & 1 \\ 
Thailand & 4 & 2 & 0 & 2 \\ 
Indonesia & 4 & 3 & 0 & 1 \\ 
Armenia & 3 & 1 & 2 & 0 \\ 
Kyrgyz Republic & 3 & 3 & 0 & 0 \\ 
Libya & 3 & 3 & 0 & 0 \\ 
Tunisia & 3 & 3 & 0 & 0 \\ 
Malta & 3 & 3 & 0 & 0 \\ 
Taiwan & 3 & 3 & 0 & 0 \\ 
Tokelau & 2 & 2 & 0 & 0 \\ 
Finland & 2 & 2 & 0 & 0 \\ 
Poland & 2 & 2 & 0 & 0 \\ 
Anguilla & 2 & 2 & 0 & 0 \\ 
Afghanistan & 2 & 1 & 1 & 0 \\ 
Uzbekistan & 2 & 1 & 0 & 1 \\ 
\end{tabular}
\end{table}

\begin{table}[h!]
\caption{\textbf{Country distribution.} Article frequency for the bottom 101-125 countries represented in the AskNews-NER-v0 dataset. Continued from Supplementary Tables \ref{tab:countries-50} and \ref{tab:countries-100}.}
\label{tab:countries-125}
\centering
\begin{tabular}{c|cccc}
\toprule
Country & Full dataset & Train split & Validation split & Test split \\ \midrule
Bahrain & 2 & 1 & 0 & 1 \\ 
Angola & 2 & 0 & 1 & 1 \\ 
Romania & 1 & 0 & 1 & 0 \\ 
Mauritania & 1 & 1 & 0 & 0 \\ 
Georgia & 1 & 1 & 0 & 0 \\ 
Benin & 1 & 1 & 0 & 0 \\ 
Hungary & 1 & 1 & 0 & 0 \\ 
British Indian Ocean Territory & 1 & 1 & 0 & 0 \\ 
St. Lucia & 1 & 1 & 0 & 0 \\ 
Fiji & 1 & 1 & 0 & 0 \\ 
Sudan & 1 & 1 & 0 & 0 \\ 
Mozambique & 1 & 1 & 0 & 0 \\ 
Greece & 1 & 1 & 0 & 0 \\ 
Uganda & 1 & 1 & 0 & 0 \\ 
Nepal & 1 & 1 & 0 & 0 \\ 
Sao Tome and Principe & 1 & 0 & 1 & 0 \\ 
Laos & 1 & 1 & 0 & 0 \\ 
Reunion & 1 & 1 & 0 & 0 \\ 
Sint Maarten & 1 & 1 & 0 & 0 \\ 
Burundi & 1 & 0 & 0 & 1 \\ 
Barbados & 1 & 0 & 0 & 1 \\ 
Lithuania & 1 & 0 & 0 & 1 \\ 
Bulgaria & 1 & 0 & 0 & 1 \\ 
San Marino & 1 & 0 & 0 & 1 \\ 
Zambia & 1 & 1 & 0 & 0 \\
\midrule
Total & 5049 & 3589 & 730 & 730 \\ 
\bottomrule
\end{tabular}
\end{table}

\begin{figure}[h!]
    \centering
    \includegraphics[width=0.95\textwidth]{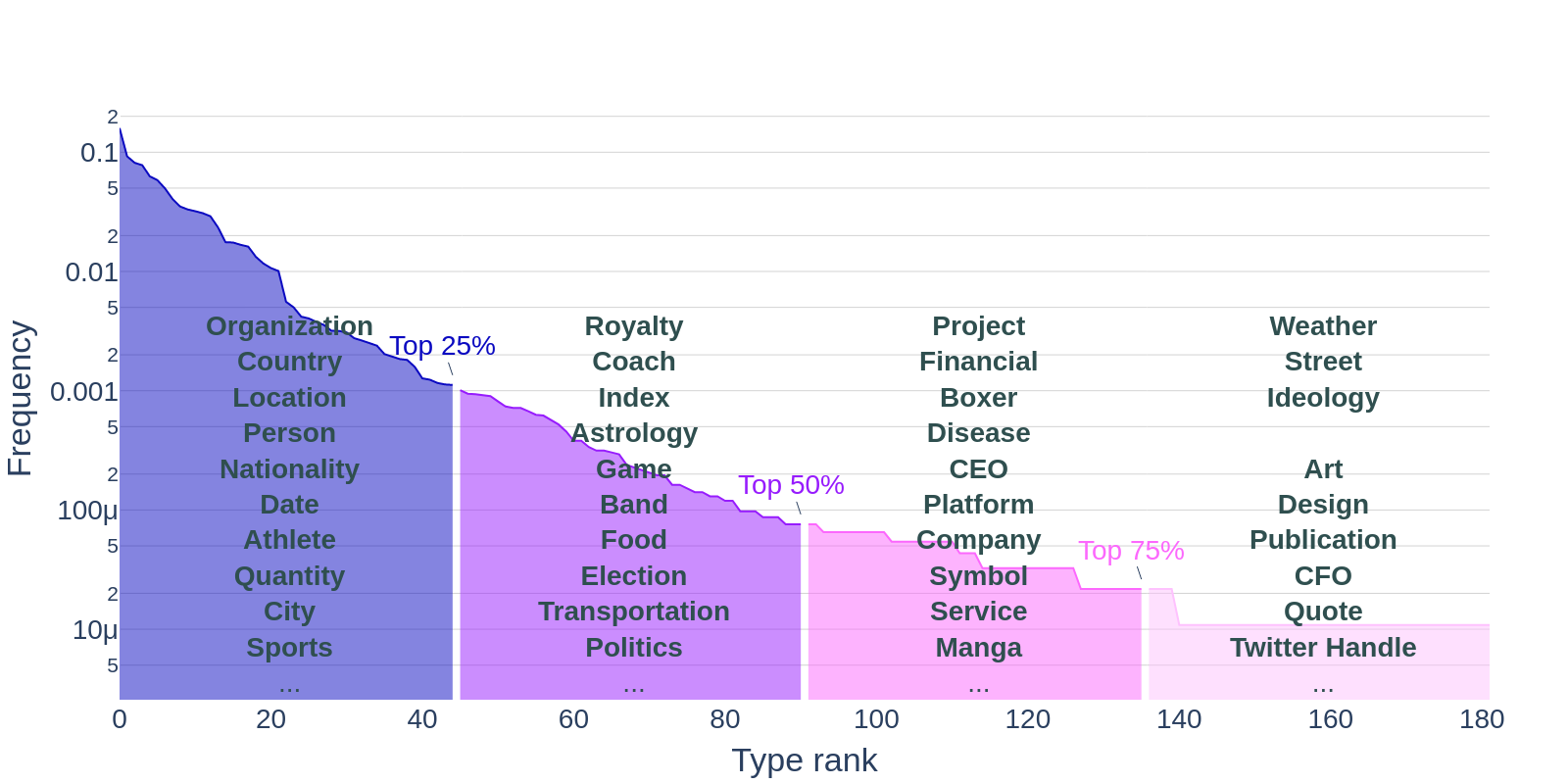}
    \caption{\textbf{Entity type frequency for the raw labels.} Labeling of news articles using Llama3 resulted in 182 unique entity types, despite requesting 54 as per Figure \ref{box:types}. Note that one of the entities types in the final 25\% of the raw labels is an empty string.}
    \label{sup-fig:types-unlimited}
\end{figure}

\clearpage
\subsection{Evaluation}

The fine-tuned NER models were evaluated in a zero-shot context on common NER benchmarks. Supplementary Table \ref{sup-tab:eval-20-all} shows the scores for GLiNER v1 models, GLiNER v2.1 models, and the fine-tuned models on 18 NER benchmark datasets. Performance improvements for the fine-tuned models compared to the base models (GLiNER v2.1) are found in Table \ref{tab:eval-20}.

\begin{table}[h!]
\caption{\textbf{Zero-shot performance on 18 NER datasets.} Comparison between base models (S - GLiNER-Small, M - GLiNER-Medium, L - GLiNER-Large) and fine-tuned models ("News"). Performance on the test split of the anNER-v0 dataset is shown at the bottom. \textbf{Bold} indicates the highest score per dataset. Commit hashes for all models are found in Supplementary Table \ref{sup-tab:commits}.}
\label{sup-tab:eval-20-all}
\resizebox{\textwidth}{!}{%
\centering
\begin{tabular}{c|ccc|ccc|ccc}
\toprule
Dataset & S-v1 & S-v2.1 & S-News & M-v1 & M-v2.1 & M-News & L-v1 & L-v2.1 & L-News \\ \midrule
ACE 2005 & 23.8 & 32.8 & 23.2 & 23.6 & 31.3 & 23.3 & 27.1 & \textbf{33.3} & 28.9 \\ 
AnatEM & \textbf{49.0} & 35.1 & 36.5 & 33.2 & 29.2 & 36.8 & 33.7 & 24.7 & 34.8 \\ 
Broad Tweet Corpus & 60.8 & 63.0 & \textbf{69.2} & 63.1 & 63.9 & 67.7 & 64.1 & 59.9 & \textbf{69.2} \\ 
CoNLL 2003 & 61.4 & 59.8 & 60.9 & 62.8 & 61.5 & \textbf{63.0} & 60.7 & 59.5 & 56.4 \\ 
FabNER & 21.8 & 17.9 & 18.9 & 24.1 & 17.0 & 20.6 & \textbf{25.4} & 19.2 & 22.5 \\ 
FindVehicle & 32.3 & 38.0 & 43.1 & 30.3 & 36.0 & 38.7 & 45.1 & 51.9 & \textbf{52.9} \\ 
GENIA\_NER & 51.2 & 47.7 & 47.2 & 57.3 & 55.4 & 54.0 & \textbf{58.4} & 55.3 & 55.0 \\ 
HarveyNER & 22.8 & 18.3 & 23.1 & 19.4 & 23.2 & \textbf{26.7} & 16.9 & 18.7 & 15.8 \\ 
MultiNERD & 58.2 & 57.3 & \textbf{68.3} & 54.9 & 55.3 & 67.1 & 54.8 & 48.7 & 64.0 \\ 
Ontonotes & 27.6 & 28.1 & \textbf{35.1} & 24.5 & 25.7 & 33.2 & 27.0 & 19.5 & 32.0 \\ 
PolyglotNER & 40.6 & 40.4 & 43.9 & 42.0 & 42.1 & \textbf{45.3} & 40.7 & 39.6 & 42.5 \\ 
TweetNER7 & 39.3 & 36.5 & 35.6 & 39.9 & 38.2 & 38.6 & \textbf{40.6} & 36.1 & 35.5 \\ 
WikiANN en & 56.4 & 55.3 & 53.6 & \textbf{58.4} & 55.5 & 53.0 & 57.4 & 55.5 & 52.5 \\ 
WikiNeural & 68.7 & 64.7 & 76.0 & 71.0 & 67.5 & \textbf{77.9} & 67.7 & 62.9 & 70.3 \\ 
bc2gm & 51.5 & 50.4 & 49.4 & 52.9 & 54.0 & \textbf{54.6} & 52.0 & 45.2 & 46.9 \\ 
bc4chemd & 44.2 & 49.0 & 49.4 & 46.4 & 47.3 & 51.1 & 48.4 & 53.1 & \textbf{55.1} \\ 
bc5cdr & 66.8 & 66.1 & 70.9 & 66.5 & 67.0 & \textbf{72.5} & 67.8 & 68.4 & 71.2 \\ 
ncbi & 55.0 & 56.6 & 57.9 & 58.6 & 60.3 & 64.5 & 63.8 & 59.7 & \textbf{65.5} \\ 
\midrule
Average & 46.2 & 45.4 & 47.9 & 46.0 & 46.1 & \textbf{49.4} & 47.3 & 45.1 & 48.4 \\ 
\midrule
\midrule
 AskNews-NER-v0 &  55.4 &  54.4 &  71.8 &  57.0 &  53.3 &  71.9 &  57.0 &  48.6 &  \textbf{72.9} \\ 
\bottomrule
\end{tabular}}
\end{table}

\clearpage
\subsection{Model versions}

For reproducibility purposes, we provide the Huggingface commit hashes for the models used in this paper (Supplementary Table \ref{sup-tab:commits}). The GLiNER models were refactored from the versions published in \citet{zaratiana2023gliner}  before being published to Huggingface. As we have evaluated our fine-tuned models compared to the versions available on Huggingface, we also compare benchmark scores of the GLiNER models available on Huggingface to those reported in in Supplementary Tables \ref{sup-tab:hf-20} and \ref{sup-tab:hf-ood}.

\begin{table}[h!]
    \caption{\textbf{Model versions.} Huggingface \includegraphics[scale=0.06]{figures/huggingface-logo.png} commit hashes for the models used for fine-tuning and evaluation in this paper.}
    \label{sup-tab:commits}
    \centering
    \begin{tabular}{l|l}
    \toprule
        Model & Commit hash \\
    \midrule
        Small-v1 & 0f0f4e7d3f10e48844110162d6b5c6072ddd5a4e \\
        Small-v2.1 & 4e091416cf7c3481db542c2a3d26156916f3a47f \\
        Medium-v1 & b63645e3596b70a7d66e0597b3e739ad17bffe9d \\
        Medium-v2.1 & a3f776a34b65ed1218fb4765b89d564260f0dc01 \\
        Large-v1 & 1f55b526b24c7576857d4eb2b047cc77b0143594 \\
        Large-v2.1 & abd49a1f1ebc12af1be84d06f6848221cf96dcad \\
    \midrule
        NuNERZero\_span & 6b5d591b348c8775ce065737e267b4aa6a907538 \\
    \midrule
        Small-AskNews-NER-v0 & dba3032b4f4426c827fed54fc9993aaece5dbf2a \\
        Medium-AskNews-NER-v0 & 44c9b148230c65df75661c38c4fa7c1825892133 \\
        Large-AskNews-NER-v0 & 8f4f822f21b393b6eb577f4efd8558894b4e2b62 \\
        \bottomrule
    \end{tabular}
\end{table}

\begin{table}[h!]
\centering
\caption{\textbf{Comparison between models currently available on Huggingface \includegraphics[scale=0.06]{figures/huggingface-logo.png} and scores reported in Table 1 in \citet{zaratiana2023gliner}.} Zero-shot performance on Out-of-Domain NER benchmark datasets for the different model sizes (S- small, M - medium, L - large). \textbf{Bold} indicates the highest score per dataset. \textcolor{green}{Green} and \textcolor{red}{red} values indicate increase or decrease in score for \includegraphics[scale=0.06]{figures/huggingface-logo.png} models compared to published scores. Commit hashes for the \includegraphics[scale=0.06]{figures/huggingface-logo.png} models are found in Supplementary Table \ref{sup-tab:commits}.}
\label{sup-tab:hf-ood}
\resizebox{\textwidth}{!}{%
\begin{tabular}{l|lllllll|l}
\toprule
Model & AI & Literature & Music & Politics & Science & Movie & Restaurant & Average \\ \midrule
S-v1 & 50.7 & 60.0 & 60.9 & 61.5 & 55.6 & 46.9 & 33.3 & 52.7 \\ 
S-v1 \includegraphics[scale=0.06]{figures/huggingface-logo.png} & 49.8 \tiny{\textcolor{red}{-0.9}} & 62.8 \tiny{\textcolor{green}{+2.8}} & 66.2 \tiny{\textcolor{green}{+5.3}} & 64.0 \tiny{\textcolor{green}{+2.5}} & 57.0 \tiny{\textcolor{green}{+1.4}} & 43.1 \tiny{\textcolor{red}{-3.8}} & 28.7 \tiny{\textcolor{red}{-4.6}} & 53.1 \tiny{\textcolor{green}{+0.4}} \\
M-v1 & 51.8 & 59.7 & 69.4 & 68.6 & 58.1 & 42.9 & 37.3 & 55.4 \\ 
M-v1 \includegraphics[scale=0.06]{figures/huggingface-logo.png} & 51.9 \tiny{\textcolor{green}{+0.1}} & 59.5 \tiny{\textcolor{red}{-0.2}} & 66.5 \tiny{\textcolor{red}{-2.9}} & 67.3 \tiny{\textcolor{red}{-1.3}} & 55.9 \tiny{\textcolor{red}{-2.2}} & 46.8 \tiny{\textcolor{green}{+3.9}} & 34.4 \tiny{\textcolor{red}{-2.9}} & 54.6 \tiny{\textcolor{red}{-0.8}} \\ 
L-v1 & \textbf{57.2} & \textbf{64.4} & \textbf{69.6} & \textbf{72.6} & 62.6 & \textbf{57.2} & \textbf{42.9} & \textbf{60.9} \\  
L-v1 \includegraphics[scale=0.06]{figures/huggingface-logo.png} & \textbf{57.2} \tiny{\textcolor{green}{+0.0}} & 63.7 \tiny{\textcolor{red}{-0.7}} & \textbf{69.6} \tiny{\textcolor{green}{+0.0}} & 72.5 \tiny{\textcolor{red}{-0.1}} & \textbf{65.3} \tiny{\textcolor{green}{+2.7}} & 48.6 \tiny{\textcolor{red}{-8.6}} & 37.6 \tiny{\textcolor{red}{-5.3}} & 59.2 \tiny{\textcolor{red}{-1.7}} \\ 
\bottomrule
\end{tabular}}
\end{table}

\begin{table}[h!]
\caption{\textbf{Comparison between large model currently available on Huggingface \includegraphics[scale=0.06]{figures/huggingface-logo.png} and scores reported in Table 2 in \citet{zaratiana2023gliner}.} Zero-shot performance on 20 NER benchmark datasets for Large models. \textbf{Bold} indicates the highest score per dataset. \textcolor{green}{Green} and \textcolor{red}{red} values indicate increase or decrease in score for \includegraphics[scale=0.06]{figures/huggingface-logo.png} models compared to published scores. The commit hash for the \includegraphics[scale=0.06]{figures/huggingface-logo.png} model is found in Supplementary Table \ref{sup-tab:commits}.}
\label{sup-tab:hf-20}
\centering
\begin{tabular}{c|cc}
\toprule
Dataset & Large-v1 & Large-v1 \includegraphics[scale=0.06]{figures/huggingface-logo.png} \\ \midrule
ACE 2005 & \textbf{27.3} & 27.1 \tiny{\textcolor{red}{-0.2}} \\ 
AnatEM & \textbf{33.7} & \textbf{33.7} \tiny{\textcolor{green}{+0.0}} \\ 
Broad Tweet Corpus & 61.2 & \textbf{64.1} \tiny{\textcolor{green}{+2.9}} \\ 
CoNLL 2003 & \textbf{64.6} & 60.7 \tiny{\textcolor{red}{-3.9}} \\ 
FabNER & 23.6 & \textbf{25.4} \tiny{\textcolor{green}{+1.8}} \\ 
FindVehicle & 41.9 & \textbf{45.1} \tiny{\textcolor{green}{+3.2}} \\ 
GENIA\_NER & 55.5 & \textbf{58.4} \tiny{\textcolor{green}{+2.9}} \\ 
HarveyNER & \textbf{22.7} & 16.9 \tiny{\textcolor{red}{-5.8}} \\ 
MultiNERD & \textbf{59.7} & 54.8 \tiny{\textcolor{red}{-4.9}} \\ 
Ontonotes & \textbf{32.2} & 27.0 \tiny{\textcolor{red}{-5.2}} \\ 
PolyglotNER & \textbf{42.9} & 40.7 \tiny{\textcolor{red}{-2.2}} \\ 
TweetNER7 & \textbf{41.4} & 40.6 \tiny{\textcolor{red}{-0.8}} \\ 
WikiANN en & \textbf{58.9} & 57.4 \tiny{\textcolor{red}{-1.5}} \\ 
WikiNeural & \textbf{71.8} & 67.7 \tiny{\textcolor{red}{-4.1}} \\ 
bc2gm & 47.9 & \textbf{52.0} \tiny{\textcolor{green}{+4.1}} \\ 
bc4chemd & 43.1 & \textbf{48.4} \tiny{\textcolor{green}{+5.3}} \\ 
bc5cdr & 66.4 & \textbf{67.8} \tiny{\textcolor{green}{+1.4}} \\ 
ncbi & 61.9 & \textbf{63.8} \tiny{\textcolor{green}{+1.9}} \\ 
\midrule
Average & \textbf{47.6} & 47.3 \tiny{\textcolor{red}{-0.3}} \\ 
\bottomrule
\end{tabular}
\end{table}


\end{document}